# Fuzzy Logic Function as a Post-hoc Explanator of the Nonlinear Classifier

Martin Klimo[1], Ľubomír Králik[1]

[1]University of Žilina, Žilina, Slovakia
{martin.klimo, lubomir.kralik}@fri.uniza.sk

**Abstract.** Pattern recognition systems implemented using deep neural networks achieve better results than linear models. However, their drawback is the black box property. This property means that one with no experience utilising nonlinear systems may need help understanding the outcome of the decision. Such a solution is unacceptable to the user responsible for the final decision. He must not only believe in the decision but also understand it. Therefore, recognisers must have an architecture that allows interpreters to interpret the findings. The idea of post-hoc explainable classifiers is to design an interpretable classifier parallel to the black box classifier, giving the same decisions as the black box classifier. This paper shows that the explainable classifier completes matching classification decisions with the black box classifier on the MNIST and FashionMNIST databases if Zadeh's fuzzy logic function forms the classifier and DeconvNet importance gives the truth values. Since the other tested significance measures achieved lower performance than DeconvNet, it is the optimal transformation of the feature values to their truth values as inputs to the fuzzy logic function for the databases and recogniser architecture used.

## 1 Introduction

Nonlinear pattern recognition systems implemented by Deep Neural Networks (DNN) reached superior performance compared to optimal linear systems [1]. Humans have experience in small, low-speed changes that we can easily approximate by linear behaviour. Therefore, we can explain the behaviour of systems under these conditions. Due to missing superposition properties, we cannot give accurate forecasts, even omitting other nonlinear systems properties like chaotic behaviour, many attractors or sensitivity to initial conditions [2]. This nonlinearity inside the black box does not allow us to explain how the classifier has reached its decision. Suppose the final responsibility for the consequences of the decision lies with the user of the neural network. In that case, the missing explanation will prevent the user from blindly using the reached decision. Therefore, decision explainability is essential in DNN research [3]. In this paper, we focus on the explainability of the classifier. We let the feature extractor find the best features to obtain more accurate recognition without dealing with their explanation.

There are two approaches to getting an explainable classification. Firstly (explainability by design) – restrict the classifier to self-explaining systems (e.g., fuzzy logical function). Secondly (post-hoc explanation) – leave the nonlinear classifier as a black box. Nevertheless, we have put together an explanation classifier to explain the decision of the black box classifier. We use this method in the paper to maintain a high recognition accuracy and utilise a fuzzy logic function as a post-hoc explainable classifier. The fuzzy logic function assumes the truth values of the logic variables at the input and generates the truth value of the logic function at the output. The primary purpose of this paper is to investigate the explainability of a black box classifier when we apply the significance of the features instead of feature values as the truth values.

Mining the importance of internal DNN variables (e.g. inputs or features) is nowadays the primary tool for explaining the decision taken [4]. However, we only consider these input variables that need support with a logical expression that will generate the same conclusion as given by the black box classifier to maintain its accuracy. More specifically, these measures of local importance are fed into a model that explains the trained classifier formed by the neural network. After normalising the importance measure, we can interpret this as a measure of the statement's truth that the feature's occurrence is necessary for the resulting decision (a negated statement is that the feature must not occur for a given decision). We must avoid the influence of features that are indifferent to the outcome of the decision (a truth value of about one-half). We label such features irrelevant and do not use them as input to an explainable classifier. The problem addressed in this paper is finding the measures of the feature importance that will be the best inputs to a fuzzy logic function in the post-hoc explainable classifier role. A similar problem is addressed in the paper [5]. L1 regularised logistic regression, and Gini importance computes the feature importance measure. Random forest and LIME represent explainable classifiers.

The remainder of the text is organised as follows: Section 2 provides the background for the proposed solution. Section 2.1 describes analysed measures of the feature importance, and Section 2.2 explains the functionality of the fuzzy logic function in the role of a post-hoc explanator. Section 3 presents the obtained results, and Section 4 concludes the paper.

## 2      Background

Figure 1 gives the principal block diagram of the post-hoc explanator. At the input of the DNN, we will attach an image sample from which the features are extracted. The nonlinear classifier recognises the class of the sample. During the training, we optimise the weights of the DNN to reach maximum precision according to the available classes in the training dataset. It is a standard procedure for training the black box recogniser.

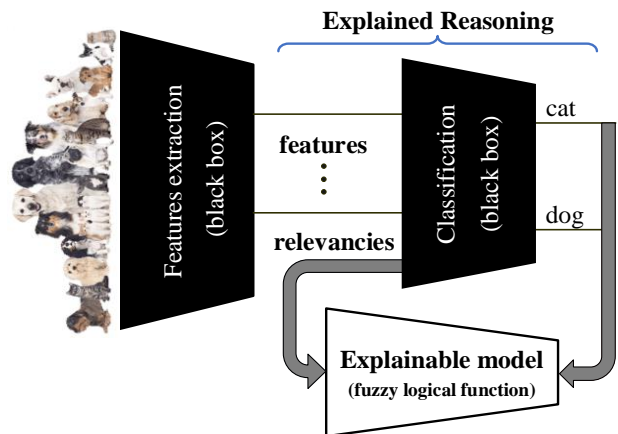

**Fig. 1.** The principal block diagram of the post-hoc explanator

To explain the black box classification, an explainable classifier is added in parallel to the black box. The logical inference is based on modus ponens IF-THEN rules forming a logic function in a disjunctive normal form. Due to the vagueness of input data, fuzzy logic was applied. We need to estimate the truth values of the features entering the fuzzy logic function to classify the input sample into possible classes. Usually, triangular or trapezoidal functions are used. The paper aims to replace these functions with more general functions mirroring features essential for the black box classifier interpretability. Methods described in Section 2.1 mine the feature importance measures and move them to the explainable classifier input. Generating truth values for the sample membership of each class and the explanation of the decision of the black box classifier is described in Section 2.2. The role of the post-hoc explainer is to mimic as closely as possible the decision-making of the black box classifier to maximise fidelity. Therefore, the paper evaluates various feature importance measures according to matching the recognised classes regardless of whether determining the categories is correct. The task of assessing the correctness of pattern recognition is left to the confidence value assessment of the black box output.

## 2.1 Feature importance measures

The input data must pass through the whole architecture with learned weights and, through nonlinear activation functions to make predictions with neural networks. A single prediction involves millions of operations. Thus, humans cannot follow an accurate mapping of the input information through the network. We must consider millions of complex parameters interacting with the data for such a process. For this reason, specific explanatory methods must interpret model variables [6].

In general, they can be broadly divided into two different types:
- Occlusion-based or Perturbation-based - methods such as LIME manipulate parts of the image to generate explanations (model-agnostic approach)
- Gradient-based - methods compute a gradient of prediction concerning input attributes.

The standard model-agnostic methods are possible to use, but methods developed explicitly for DNNs have some advantages, such as:
- DNNs learn features and concepts in their hidden layers; hence unique methods for their interpretation are needed.
- The gradient could be used in the implementation of explainable methods, thus making the computation more efficient than model-agnostic methods.

The most important for our purpose is Gradient-based methods because they carry the information about how much a slight change in image pixel would affect the model's prediction. Explainable gradient-based methods are, therefore, a specific case of feature assignment, or more precisely, in this case, pixel assignment, since pixels, words, or tabular data can represent the features. These methods explain individual model predictions by assigning each input feature to how much it affected the prediction (negative or positive).

These approaches usually produce the input image interpretation. Each pixel is assigned a value representing its importance for the final prediction or classification. This property can be extended into any feature vector produced by the DNN. Thus, we can create the vector containing the relevance measures that we can use later as the input to the post-hoc fuzzy explanator.

Saliency Map [7] is the first and most straightforward method for feature importance measures. This method computes the gradient of the selected neuron we want to explain concerning the individual pixels of the input image or, in general, concerning features we want to interpret.

The following considered method DeconvNet [8] slightly modifies the previous method in terms of changing the computation of the gradient. In this case, only positive gradient values are propagated, and any negative ones are simply set to zero.

The third method, Guided backpropagation [9], combines the abovementioned methods into a single one. Such an approach can be viewed as a strict filtering rule that keeps only the most essential gradient values.

The last method considered in this paper for feature importance measures is called Layer-Wise Relevance Propagation (LRP) [10, 11], which is a method for interpreting the decisions made by DNN by assigning relevance scores to the features accessing the classifier. The method propagates the relevance scores from the output layer back through the network, layer by layer, using a set of propagation rules. These rules ensure that the relevance scores are conserved, meaning that the total relevance at the output of a layer is equal to the total relevance at its input.

## 2.2 Post-hoc fuzzy explanator

The main task of the post-hoc explainable classifier is to give the decision regarding the recognised class in the form of a logical statement. The paper [12] assigns a binary code vector to each feature vector. The bits of this vector may point to features that must be (bit equal to one) and not be (bit equal to zero) present in the sample. However, some features are irrelevant to the placement of the sample into the correct class. We call these features "irrelevant" and assign them the value "X" in the codeword. The codeword bits express the qualitative impact of the features on the decision and allow the IF-THEN logical rules to be written. While the input samples are vague, we must introduce the fuzzy statements' truth values. It concerns the decision statement "the class of the sample P is A" and the input statements like "the feature B must be present". To exclude the irrelevant features, we do not assign truth values to them; they are excluded from evaluating the statements by a relevance filter. We have to set a fuzzy equality of necessity of occurrence for the relevant features – the truth value from the unit interval. Notation $t(feature = 1)$ means the truth degree of the statement "the feature must be present". Assignment of the truth values to the input variables (features in our case) usually follows the shapes of the membership functions in fuzzy sets. In most applications, they have triangular or trapezoidal forms. This paper studies whether other forms of the truth function give a better fit between the predictions of a logic function and a black box classifier. Because the interpretability of features evaluates their impact on the recognition result, we derive the truth value measure of features from their relevance. Since the relevance is a real number while the truth value is from the unit interval, the first step is *min-max* linear normalisation

$$\tilde{y}_i = \frac{y_i - y_{min}}{y_{max} - y_{min}}, \quad y_{min} = min\{y_1, \dots, y_M\}, \quad y_{max} = max\{y_1, \dots, y_M\}, \quad (6)$$

where $y_i \in \mathcal{R}$, $i \in \{1, \dots, M\}$ are the features importance measures, and $\tilde{y}_i \in [0,1]$, $i \in \{1, \dots, M\}$ are the features truth values. The second step is mapping importance into relevance categories: c=0 is a negative relevance, c=1 is a positive relevance, and c=X is an irrelevant feature. Rounding

$$c_i = \begin{cases} 1, & \tilde{y}_i > \frac{1}{2} + \Delta \\ X, & \frac{1}{2} - \Delta \leq \tilde{y}_i \leq \frac{1}{2} + \Delta, \quad i \in \{1, \dots, M\}, \\ 0, & \tilde{y}_i < \frac{1}{2} - \Delta \end{cases} \quad (7)$$

(where $\Delta$ gives an irrelevance range), assigns the relevance codewords from the set

$$C_{full} = \{c^j = (c_1^j, \dots, c_M^j) \in \{0, X, 1\}^M\}, j = 1, \dots, 3^M$$

to each truth vector $\tilde{y} = (\tilde{y}_1, \dots, \tilde{y}_M) \in [0,1]^M$. $c_i^j$ means the i-th coordinate of the j-th codeword. After filtering irrelevant features, we calculate a truth value of the statement that the feature is relevant (negatively or positively)

$$t(\tilde{y}_i = c_i) = \begin{cases} 1 - \tilde{y}_i, & c_i = 0 \\ \tilde{y}_i, & c_i = 1 \end{cases}, c_i \neq X, \quad i \in \{1, \dots, M\}, \quad (8)$$

and how relevant the codeword is

$$t(\tilde{c} = c^j) = \min_{\substack{i=1,\dots,M \\ c_i^j \neq X}} t(\tilde{c}_i = c_i^j), j = 1, \dots, 3^M. \quad (9)$$

A "raw data" example (with a linear truth value function) is also tested to evaluate the truth value optimisation concept. In this case, the feature values themselves replaced the feature relevancies.

As one can see, we apply the Zadeh fuzzy logic. Figure 2 shows the block diagram of an explainable classifier based on the fuzzy logic function. Omitting irrelevant features ensures that they do not affect the truth value of the fuzzy logic statements.

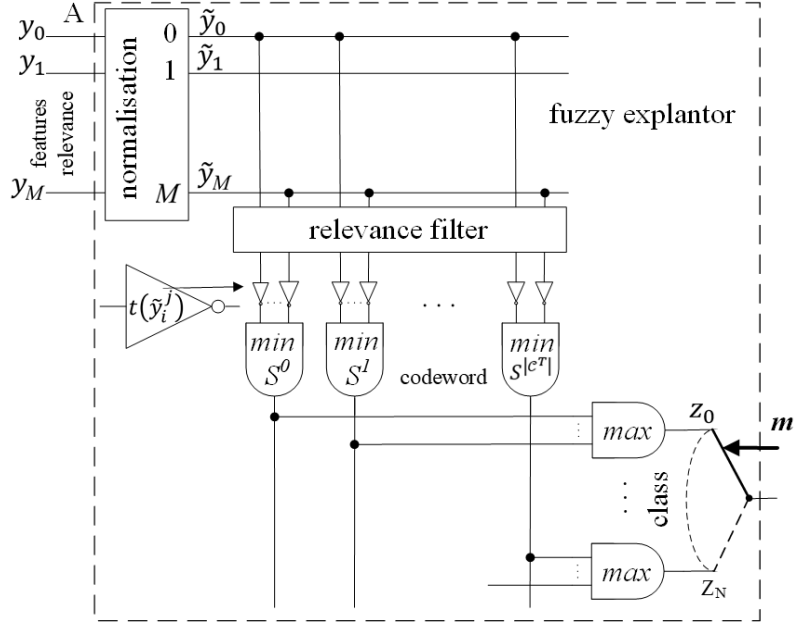

**Fig. 3.** The fuzzy logical explanator

The result of training is the set of codewords $C^T = \{C_1^T, ..., C_N^T\} \subset C_{full}$ that occurred during the training, where the codeword $c \in C_i^T$ belongs to the class $i \in \{1, ..., N\}$. Let the black box forecast $\widetilde{m} \in \{1, ..., N\}$ as a winning class, and the winning class from the explainable classifier is the class $\widetilde{\widetilde{m}} \in \{1, ..., N\}$ associated with the codeword

$$\tilde{c} = \underset{j:c^j \in C^T}{\mathrm{argmax}} \min_{\substack{i=1,...,M \\ c_i^j \neq X}} t(\tilde{c}_i = c_i^j). \tag{10}$$

We can look at the explainable and the black box classifiers as competing systems and ask which one has higher classification accuracy. However, the main task of the explainable classifier is to explain the decisions of the black box classifier. Therefore, we will consider it a success if its classification joins the black box classifier and the success rate is

$$p = \frac{\sum_{i=1}^{n} \delta(\widetilde{\widetilde{m}}_i = \widetilde{m}_i)}{n}, \tag{11}$$

where $\delta(\widetilde{\widetilde{m}}_i = \widetilde{m}_i) = \begin{cases} 1, \widetilde{\widetilde{m}}_i = \widetilde{m}_i \\ 0, \widetilde{\widetilde{m}}_i \neq \widetilde{m}_i \end{cases}$ and $n$ is the number of tested samples. We evaluate in Section 3 several measures of the feature importance according to this measure.

To explain the decision of the black box classifier, we identify the codeword with the highest truth value within the codewords belonging to the same class $C_{\widetilde{m}}^{MT}$ within the explainer

$$\tilde{c} = \underset{j:c^j \in C_{\widetilde{m}}^T}{\mathrm{argmax}} \min_{\substack{i=1,...,M \\ c_i^j \neq X}} t(\tilde{c}_i = c_i^j). \tag{12}$$

We can display this relevance codeword $\tilde{c} = (\tilde{c}_1, \tilde{c}_M), \tilde{c}_i \in \{0, X, 1\}$, its truth value $t(\tilde{c}) = \min_{\substack{i=1,...,M \\ \tilde{c}_i \neq X}} t(\tilde{c}_i)$ and truth values of the codeword components $t(\tilde{c}_i)$, $\tilde{c}_i \neq X$. Available are also features values, their importance measures $y_i \in \mathcal{R}$ or normalised importance measures $\tilde{y}_i \in [0,1]$.

# 3 Results

For designed experiments, we used two different datasets on the same architecture. By doing so, we can compare other feature extraction methods and their effectiveness. The first dataset is the well-known MNIST [13] which contains Handwritten Digits images. The second one – Fashion MNIST [14], is a slightly challenging version of the Base MNIST. The images are still grey-scaled, with exact resolution and quantity. However, this data consists of different clothing with more complex shapes. We trained LeNet-5 [13] architecture on these datasets, but we performed enhancements to reach this model's state-of-the-art accuracy. The convolution part of the net was expanded, aiming at more complex features, while the classification head was denser and deeper to accommodate increased network parameters. The hyperparameters and regularisation were also included and tweaked for the best performance. Optimisations significantly helped push the accuracy of the model over 99%. Everything was assembled in the TensorFlow framework. Table 1 gives results obtained for MNIST and FashionMNIST datasets. The reported values represent the success rate according to (11). We have trained ten models with different starting seeds to get more accurate results. So, the reported values are averaged over these individual runs of training. The average black box's accuracy for MNIST and FashionMNIST is 99.61% and 92.46 %, respectively.

**Table 1.** Rate of matching between explainable and black box classifier

| Feature importance method | MNIST | FashionMNIST |
|---|---|---|
| Raw features | 99.20 % | 91.83 % |
| Saliency maps | 99.47 % | 98.33 % |
| **DeconvNet** | **100.00 %** | **100.00 %** |
| Guided Backpropagation | 99.79 % | 99.59 % |
| LRP | 98.35 % | 93.38 % |

We compute the feature interpretation methods from chapter 2.1 and the respective metrics to compare these methods (11). Also, to conclude if the post-hoc explanatory model can match our nonlinear classifier in terms of accuracy, thus explain it. The gradient-based methods provide highly accurate results, while DeconvNet outperformed them with perfect compliance with the classifier. The LRP method with the $\varepsilon$ rule seems to give the worst feature information but is still relevant with high accuracy. Unexpectedly, the raw normalised feature vector after feature extraction from the LeNet-5 classifier is sufficient to feed the fuzzy explanatory with the appropriate input information.

Looking at Figures 4 – 7 below can achieve better intuition behind our proposed approach. Figures 4 and 5 depict the example of accurate classification of the handwritten image of 8 (shown in Figure 4). Figures 6 and 7 depicted the example of inaccurate classification when the handwritten image of 8 was classified as the number 6. One can see that the number of Positive relevance features is relatively low compared to the number of Negative relevance features. Note that the features which do not appear in the figure have a truth value equal to zero. This value means that the test sample must indeed not contain these features. In the presented case, we used a uniform distribution of truth values over the unit interval among positively relevant, irrelevant, and negatively relevant features: $c_i = X$, $\frac{1}{2} - \Delta \leq \tilde{y}_i \leq \frac{1}{2} + \Delta$, $\Delta = \frac{1}{6}$ (truth value above 66%). This margin can be experimented with, and increasing it would reduce the number of negatively relevant features and make the decision easier to interpret. Truth values of critical features are circled in Figures 4 a) and 6 a). Unlike interpreting image pixels, when explaining a pattern, the user loses the connection between the image and the result of its recognition. Therefore, we underline that the explanation of the classification result is only a partial result that must coexist with the explanation of the relevant features. This task can be set aside for a separate study and is not part of the paper.

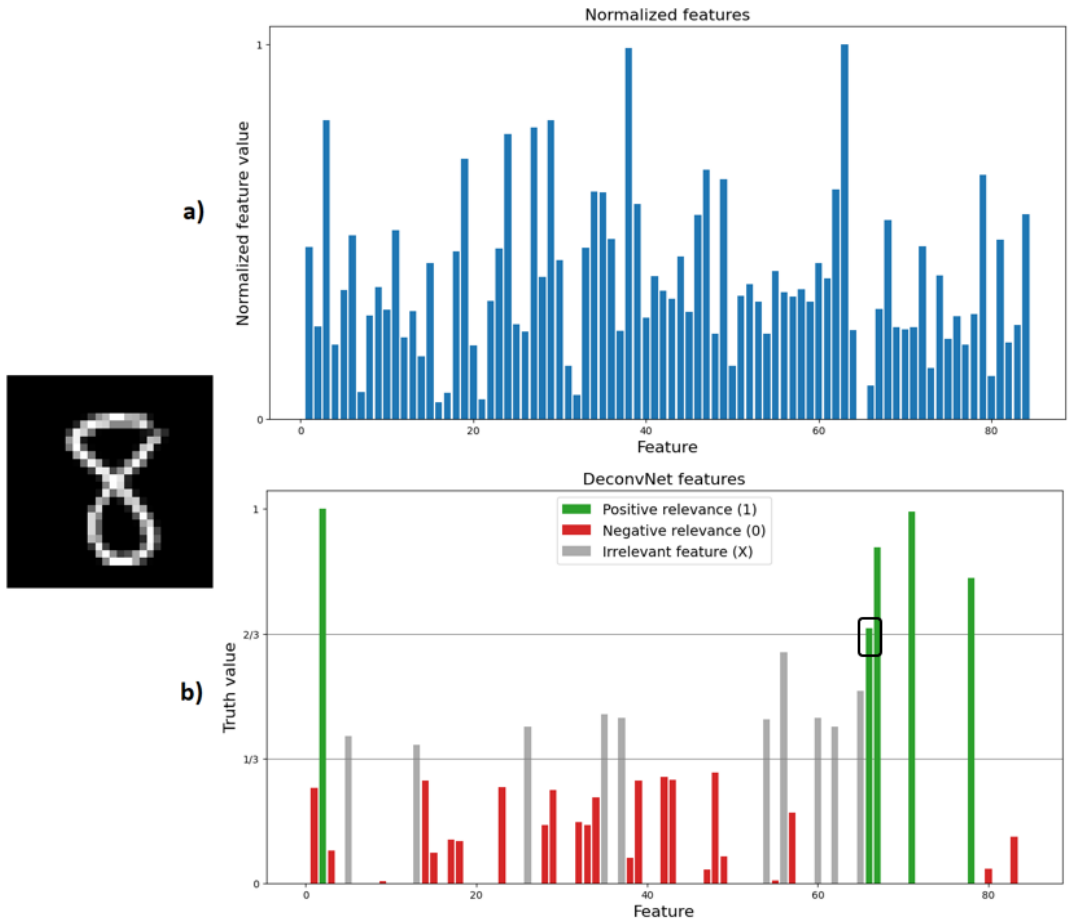

**Figure 4**: **a)** Normalised features vector from LeNet-5 layer after feature extraction on *accurate classification*, **b)** Computed features by DeconvNet method with Positive/Negative and Irrelevant contributions

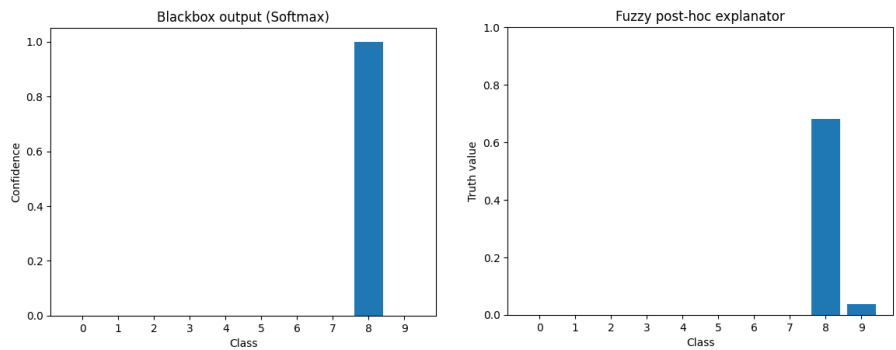

**Figure 5**: Confidence score in case of *accurate classification* given by LeNet-5 classifier (on the right) and truth value given by Fuzzy post-hoc explanator (on the left) for the input image from Fig. 1.

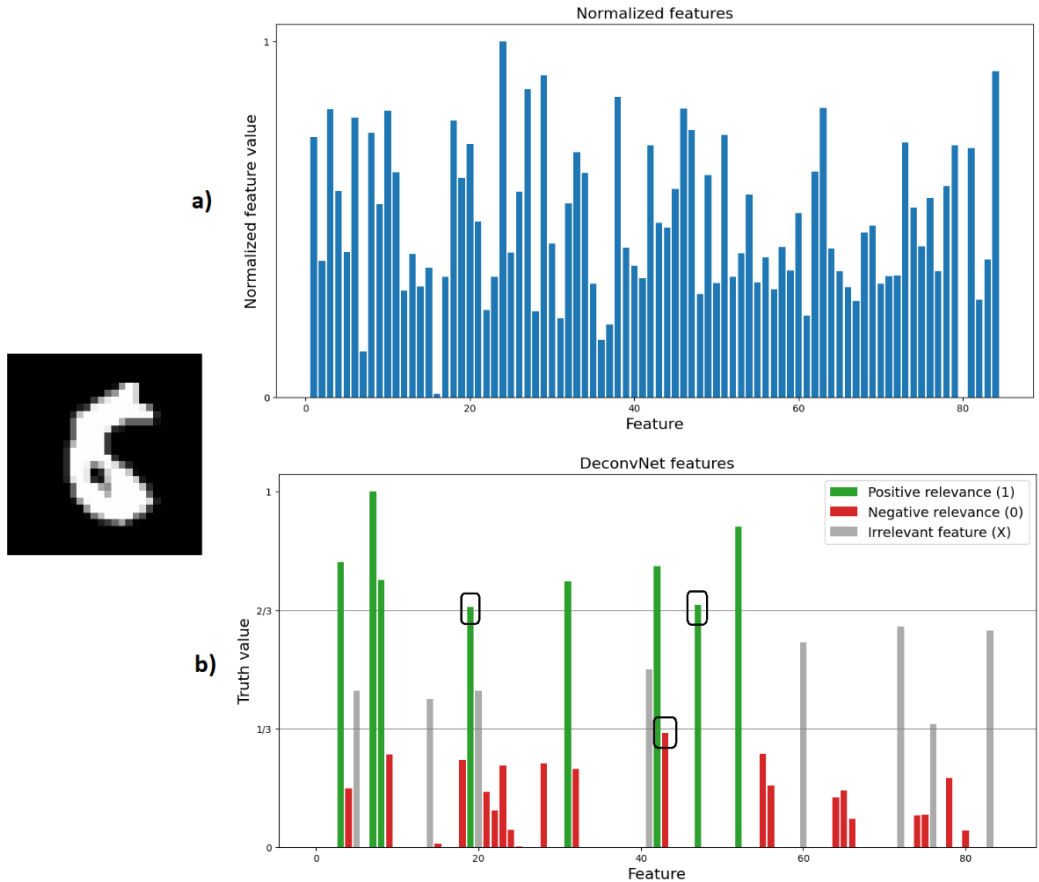

**Figure 6**: **a)** Normalised features vector from LeNet-5 layer after feature extraction on data image with *wrong classification,* **b)** Computed features by DeconvNet method with Positive/Negative and Irrelevant contributions

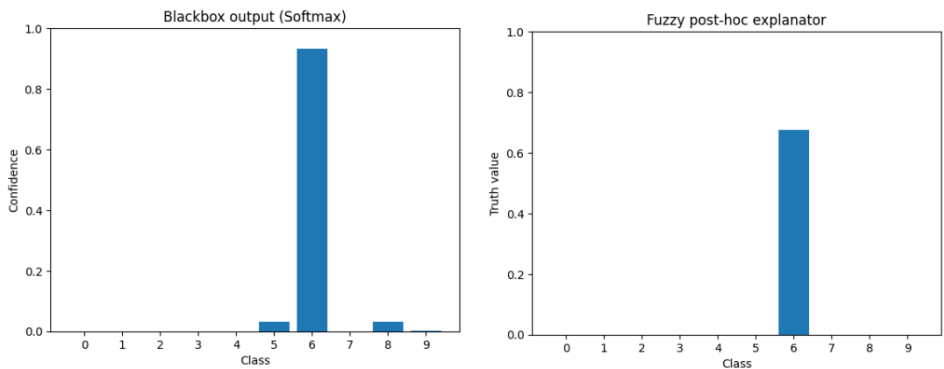

**Figure 7**: Confidence score in the case of *wrong classification* given by LeNet-5 classifier (on the right) and truth value given by Fuzzy post-hoc explanator (on the left) for the input image from Fig. 1.

## 4  Conclusions

Pattern recognition systems implemented using deep neural networks achieve better results than linear models. However, their drawback is the black box property. This property means that one with no experience utilising nonlinear systems may need help understanding the outcome of the decision. Such a solution is unacceptable to the user responsible for the final decision. He must not only believe in the decision but also understand it. Therefore, recognisers must have an architecture that allows interpreters to interpret the findings. European culture has a two-thousand-year history of using Aristotle's logic. Consequently, we assume that a decision expressed as a logical statement will be meaningful to the user.

This paper focuses on the explainability of a classification subsystem that generates a decision from the extracted features. To respect the data uncertainty, we apply Zadeh's fuzzy logic. The question is how adding an explainability condition will worsen the recognition success rate. Therefore, it is not desirable to replace the black box classifier with an explanatory classifier but to require the explanatory classifier to explain the decision of the black box classifier. Therefore, when designing a post-hoc classifier, we need it to replicate the findings of the black box classifier as closely as possible.

The inputs to the fuzzy logic function are the truth values of the features. This demand does not mean these are the feature values normalised to a unit interval, just as pixels saturation does not express their significance. Today, a palette of measures of pixel importance in an image is available, which can also be used to measure features' extent (truth value). We compared the four most widely used relevance measures by applying raw feature values: Saliency Maps, DeconvNet, Guided Backpropagation and Layer-Wise Relevance Propagation. The MNIST and FashionMNIST databases show that importance levels can be worse or better than the directly unit-interval-normalised feature values. Using the DeconvNet as the feature truth value on the fuzzy logic classifier inputs emerges as the clear winner of the above tests. In this case, the explainable classifier ultimately achieves the decisions of the black box classifier. In all instances of the test set, the best explainable decisions are those by the black box classifier. DeconvNet provides the best measure of feature importance from a scrubbing perspective. Suppose this result is confirmed on multiple databases and recognition systems. In that case, it will imply that DeconvNet is the optimal feature importance measure to maximise the fitting of the post-hoc fuzzy explainer. To conclude, the main findings are:

- The post-hoc explainer of the black box classifier based on Zadeh fuzzy logic gives the perfect fitting on MNIST and FashionMNIST databases.
- The best feature into a truth value transform among tested methods of interpreting the features given the DeconvNet.
- To exclude irrelevant features from the classification procedure, setting the truth value to one-half is not sufficient, and they should be filtered out. The feature has the X state set if the truth value belongs to the irrelevancy range. No general rule for the irrelevance range finding was not obtained yet.
- The main goal is a class prediction fitting the explainable and black box classifier. Therefore the explainable classifier cannot detect misclassification. The evaluation of confidence and truth degree is left for further research.
- A generalisation of the feature transforms into truth values is left for further research.

## Acknowledgement

This research was funded by KEGA grant number 008ŽU-4/2021.